\documentclass[journal]{IEEEtran}

\usepackage{graphicx}
\usepackage{amsmath}
\usepackage{booktabs}
\usepackage{url}
\usepackage[hidelinks]{hyperref}
\usepackage{tikz}
\usetikzlibrary{shapes.geometric, arrows.meta, positioning}

\begin{document}

\title{Caption-Mediated Perceived-Safety Estimation for Pedestrian Routing}

\author{Simon Parkinson, Paloma Liu, Wei Zheng, and Mohammadreza Sheikhfathollahi 
\thanks{Manuscript received XX XXXX 2026. This work was supported by RISE, an accelerated programme delivered by SPRITE+ with SALIENT and the Network for Security Excellence and Collaboration (NSEC) at The University of Manchester, funded by UK Research and Innovation via the Economic and Social Research Council under Grant UKRI 4192 / R133854.}
\thanks{S. Parkinson is with the School of Computing and Mathematics, Manchester Metropolitan University, Manchester M1 5GD, U.K. (e-mail: s.parkinson@mmu.ac.uk).}%
\thanks{P. Liu and M. Sheikhfathollahi are with the University of Huddersfield, Huddersfield HD1 3DH, U.K.}%
\thanks{W. Zheng is with The University of Manchester, Manchester M13 9PL, U.K.}}

\markboth{IEEE TRANSACTIONS ON INTELLIGENT TRANSPORTATION SYSTEMS}%
{Parkinson \MakeLowercase{\textit{et al.}}: Caption-Mediated Perceived-Safety Estimation for Pedestrian Routing}

\maketitle

\begin{abstract}
This paper presents an explainable approach to pedestrian routing, in which perceived safety is estimated from street-level imagery through an explicit natural-language intermediate representation. A vision--language model caption is generated and stored before any scoring is undertaken, and the perceived-risk class is derived entirely from structured features of that stored text, so that every segment score remains inspectable by the user. Nine captioning conditions across five model families are benchmarked against a direct Contrastive Language--Image Pre-training (CLIP) image-embedding baseline under an identical downstream pipeline, and the caption-mediated representation is found to reach parity with the image embedding rather than to trail it. The approach was deployed over 654,115 images covering 36 electoral wards in two locations in Northern England (Manchester and Huddersfield). Independent field validation against 3,669 locally collected ratings of 494 images across 70 participant sessions established agreement that is statistically significant but modest, at $r=0.262$, against a measured noise ceiling of 0.737 imposed by disagreement between raters. A single-use confirmatory test then found that a pipeline 44\% stronger on the supervised benchmark did not produce measurable improvement in the field ($r=0.250$, $p=0.84$), so the benchmark gains did not predict the deployment gains in this case. Routing behaviour varies systematically with journey length. There is negligible change below 1\,km, reaching a median increase of 12.78\% in low-risk route length for a median detour of 2.73\% on journeys of 3 to 6 km. 
\end{abstract}


\section{Introduction}
\label{sec:intro}
\IEEEPARstart{F}{ear} of crime exerts a measurable influence on daily mobility. More specifically, people alter routes, avoid particular streets, change travel mode, or avoid entire journeys based not only on actual risk, but on how an environment \emph{feels}~\cite{valentine1989geography,pain2001gender}. It should be noted that this influence is not evenly distributed throughout the population and is most acute where the risk of harassment or violence is highest. Violence against women and girls (VAWG) remains a high-volume and substantially under-reported category of crime, and the fear of it constrains movement well beyond those incidents that are actually recorded. \emph{Perceived safety} is therefore a useful measurement in its own right for pedestrian routing.

Recorded crime data is the dominant input to safety-aware routing systems~\cite{sohrabi2022safe,bura2019predicting}; however, it is poorly suited to this task. More specifically, it records where offences were reported rather than where people feel unsafe, and it is known to under-represent precisely those offence categories most relevant to VAWG. Manual environmental auditing provides a direct measure of the environment itself; however, it is labour intensive and does not scale to the area of a city. The use of street-level imagery offers a scalable alternative, and there is now a substantial body of work in which perceived-safety scores are derived from such imagery~\cite{salesses2013collaborative,naik2014streetscore,dubey2016deep,zhang2024urban}. Two key limitations of that literature motivate the research presented in this paper.

The first limitation concerns auditability. Most systems produce a perceptual judgement \emph{directly} from pixels or embeddings, and so the basis of a given score is difficult to interrogate. More specifically, where textual descriptions are produced, they are generated alongside the score as a post-hoc justification rather than being the representation from which the score is computed~\cite{zhang2024urban}. This distinction matters because a system that informs where people walk, or where authorities invest in environmental improvements, must allow individual judgements to be audited. The second limitation concerns validation. Such systems are typically validated against portions of the same crowdsourced dataset used during training~\cite{naik2014streetscore,dubey2016deep,zhang2024urban}.
Independent validation against human judgements collected \emph{in the deployment area} is rare, with site visits and expert ratings used in~\cite{solymosi2024using} being a notable exception, and the reliability of the human reference itself is rarely quantified (Section~\ref{sec:related-validation}). This is significant as it means that a reported agreement figure cannot be compared across studies without knowing how well the reference was measured.

In this paper, both limitations are addressed. Our aim is to provide a routing capability that is able to estimate perceived safety at city scale whilst remaining legible to the people it routes and to the local authority officers who might act upon it. We describe a pedestrian routing system in which perceived safety is estimated through an explicit caption-mediated pathway, and we report a field validation study. The work is organised around the following three research questions.

({\bf RQ1}) \textbf{Can caption mediation be used to predict perceived safety, and if so, how does it compare against alternative approaches?} Generating a perceptual judgement through natural language makes that judgement explainable to the user; however, text is a narrower channel than a learnt embedding. It is therefore important to establish how much accuracy is given up relative to classifying directly from image embeddings, and whether any deficit is attributable to the use of text, or instead to the features extracted from it. 

({\bf RQ2}) \textbf{How does a safety-weighted route compare with a route optimised for distance?} Weighting a pedestrian network by predicted safety is straightforward to implement; however, whether it produces a worthwhile exchange is not known. More specifically, what detour does the weighting incur, and what change in exposure does that detour deliver? 

({\bf RQ3}) \textbf{How valid are the predictions of the deployed model?} This question has three parts: (1) how closely does the deployed classifier agree with the judgements of people in the area it scores; (2) what upper bound does disagreement among those same people place on any such agreement; and finally, (3) does an improvement measured on the supervised benchmark translate into improved agreement in the field? 

Against these questions, this paper makes four contributions:

\begin{enumerate}
\item \textbf{A caption-mediated perceived-safety pipeline} in which a vision--language model caption is stored as an independently inspectable artefact \emph{before} any scoring, and the perceived-risk class is derived entirely from the structured features of that text (Section~\ref{sec:method}). Because that intermediate representation is model-agnostic text, nine captioning conditions across five model families become directly comparable using the same pipeline.
\item \textbf{A quantification of what caption mediation costs}, benchmarked against a CLIP image-embedding baseline under that same pipeline (Section~\ref{sec:benchmarks}). An expectation that a textual intermediate carries an accuracy penalty proves largely attributable instead to the \emph{text representation}, and a ten-dimensional probe of CLIP built from named environmental-design constructs recovers macro-F1 0.776 against the full embedding's 0.796, indicating that, for this probe, interpretability and embedding-level accuracy are not strictly opposed.
\item \textbf{A city-scale deployment} over 654,115 images across 36 electoral wards in Manchester and Huddersfield, UK, transferring between the two locations without retraining, and characterised over 3,000 sampled origin--destination pairs, stratified by journey length in both cities (Section~\ref{sec:routing-behaviour}).
\item \textbf{Field validation establishing the measurement limits in this deployment context}, comprising 3,669 locally collected ratings of 494 images across 70 participant sessions (Section~\ref{sec:validation}). Inter-rater disagreement bounds the attainable correlation at $r=0.737$, of which the classifier reaches 35\%, and a pipeline 44\% stronger on the supervised benchmark produced no measurable field improvement (Section~\ref{subsec:transfer}).
\end{enumerate}

The remainder of this paper is structured as follows. Section~\ref{sec:related} presents and discusses related work. Section~\ref{sec:method} presents and justifies the method used in this research. Section~\ref{sec:deployment} describes the two locations used in this study and the resulting deployment statistics. Section~\ref{sec:benchmarks} presents the captioning and representation benchmarks. Section~\ref{sec:routing-behaviour} then investigates how the resulting safety scores behave when used as a metric for optimising pedestrian routes. This leads to the human-rater evaluation in Section~\ref{sec:validation}, before the approach and its results are discussed in Section~\ref{sec:discussion}. Finally, a conclusion is provided in Section~\ref{sec:conclusion}, discussing key achievements, limitations, and a roadmap for future work.

\section{Related Work}
\label{sec:related}

This section is structured as follows. First, work establishing perceived safety from street-level imagery is surveyed. Second, safety-aware route planning is considered. Finally, validation practice and the reliability of human references are discussed, as this underpins the analysis presented in Section~\ref{subsec:ceiling}.

\subsection{Perceived Safety from Street-Level Imagery}
The quantification of perceived urban safety from street-level imagery was undertaken at scale in the Place Pulse project, which crowdsourced pairwise comparisons of Google Street View images across perceptual dimensions including safety~\cite{salesses2013collaborative}. In related work, Streetscore applied support vector regression over low-level image features to predict perceived safety across 21 US cities~\cite{naik2014streetscore}. This was subsequently extended using convolutional networks trained on Place Pulse 2.0, comprising 110,988 images, 1.17 million comparisons and 56 cities, thereby enabling global-scale perceptual scoring~\cite{dubey2016deep}. Related work has utilised such models to study physical urban change~\cite{naik2017computer} and to directly learn high-level perceptual judgements~\cite{ordonez2014learning}. This body of work is useful in establishing that perceptual judgements can be recovered from imagery at scale; however, the judgement itself remains opaque, which motivates the approach taken here.

More recently, multimodal large language models have been applied to the task directly, either by prompting a model to reproduce Place Pulse-style comparisons, or by retrieving against human-annotated anchors using image embeddings, with reported agreement of $R^2=0.40$--$0.45$ across two study areas~\cite{zhang2024urban}. The approach taken in this research differs in the \emph{ordering} of representation and judgement. More specifically, rather than treating text as an explanation attached to an otherwise opaque score, a natural-language caption is generated and stored first, and the safety label is then derived entirely from structured features of that caption. This makes the representation upon which the judgement rests available for inspection, which results in captioning models becoming directly comparable under a fixed downstream pipeline.

A parallel line of work represents images not by generated text, but by contrastive image--text embeddings. Contrastive Language--Image Pre-training (CLIP) aligns images and captions in a shared space through contrastive pre-training over large web corpora, and its embeddings transfer to downstream visual tasks with little or no fine-tuning~\cite{radford2021learning}. Such embeddings are frequently strong baselines; however, it is worth noting that they are not descriptions. More specifically, the model scores the correspondence between an image and candidate text and cannot produce a caption, which results in a representation that is a vector rather than an artefact a user is able to read. CLIP is therefore adopted here not as a competing system, but as a reference against which the cost of insisting on an explainable and readable caption can be measured (Section~\ref{sec:benchmarks}).

This work also builds upon a line of research that applies automated computer vision to street imagery in the context of crime and fear of crime. Dakin et al. detected eight predetermined features of the built-environment and tested their dependence on the type of crime recorded~\cite{dakin2020built}. Solymosi et al. introduced Computational Systematic Social Observation, cataloguing several hundred automatically detected features within resident-drawn safe and unsafe polygons, and validating the resulting associations against site visits and expert ratings~\cite{solymosi2024using}. Both approaches are effective in establishing reproducible associations at scale; however, both operate over discrete object vocabularies. Solymosi et al. explicitly identify the resulting loss of compositional and contextual information as a limitation, noting that this context is often precisely what a human coder relies upon. More specifically, a word set records the presence of a bench or a tree, but not the condition of that bench, nor why that particular tree matters for a sightline. The use of a free-form caption rather than a word set is a direct response to this limitation. The environmental cues of interest, namely lighting, enclosure, disorder, and natural surveillance, derive from Crime Prevention Through Environmental Design (CPTED)~\cite{monchuk2019towards,dakin2020built}.

\subsection{Safety-Aware Route Planning}
The incorporation of safety into route planning is a mature area. The majority of safety-aware route-finding approaches, for both vehicular and pedestrian travel, weight candidate routes using historical crime or crash statistics rather than a directly observed measure of the environment~\cite{sohrabi2022safe}. Within women's safety specifically, systems have been proposed that predict secure walking routes from recorded crime mapped onto a road network~\cite{bura2019predicting}. These solutions are similar in intent to those explored in this research; however, they are based on reported incidence rather than visual perception at the street level. This is significant as it means that a street without reported offences is treated as safe irrespective of the situation it presents to a pedestrian.

The routing component used in this research is a weighted search of the shortest-path~\cite{dijkstra1959note}, which compares a route only by distance against a route weighted by a safety cost per-edge. This is deliberately conventional, as the contribution of this paper lies in the derivation of the weight rather than in the search itself.

\subsection{Validation Practice and the Reliability of Human References}
\label{sec:related-validation}
In the work surveyed above, model quality is generally reported as agreeing with human judgments, expressed as classification precision, $R^2$, or a ranking correlation. The reliability of those human judgments is rarely reported in conjunction with them. This omission is significant because agreement with a noisy reference is bounded by the reliability of that reference. The relationship is formalised by classical attenuation theory and is routinely handled in psychometrics via the Spearman--Brown correction~\cite{spearman1910correlation,brown1910some,shrout1979intraclass}, and in neuroimaging and psychophysics through explicit noise-ceiling estimation~\cite{nili2014toolbox}. To the best of the authors' knowledge, ceilings of this kind have not been reported for street-view perceived-safety systems. Section~\ref{subsec:ceiling} does so.

\section{Method}
\label{sec:method}

\begin{figure}[!t]
\centering
\begin{tikzpicture}[
 node distance=0.42cm,
 every node/.style={font=\footnotesize},
 stage/.style={rectangle, rounded corners=2pt, draw=black!70, fill=black!4,
        text width=6.8cm, align=center, minimum height=0.62cm, inner sep=3pt},
 arrow/.style={-{Latex[length=1.6mm]}, thick, black!70}
]
\node[stage] (a) {\textbf{Imagery}\ \ 20\,m network sampling, one record per\\ panorama, four headings per panorama};
\node[stage, below=of a] (b) {\textbf{Caption}\ \ vision--language model $\rightarrow$ stored text\\ (9 conditions, 5 model families)};
\node[stage, below=of b] (c) {\textbf{Features}\ \ tokenisation, LDA topic memberships,\\ sentiment-derived safety score};
\node[stage, below=of c] (d) {\textbf{Class}\ \ supervised classifier $\rightarrow$ \{Low, Neutral, High\}\\ perceived risk, trained on USID fear-of-crime scores};
\node[stage, below=of d] (e) {\textbf{Edge weight}\ \ mean severity of nearby classified\\ locations $\rightarrow$ $w_e$ on pedestrian network};
\node[stage, below=of e] (f) {\textbf{Routing}\ \ Dijkstra by length, and by length$_e w_e$;\\ associated captions displayed along the route};
\draw[arrow] (a) -- (b); \draw[arrow] (b) -- (c); \draw[arrow] (c) -- (d);
\draw[arrow] (d) -- (e); \draw[arrow] (e) -- (f);
\end{tikzpicture}
\caption{The caption-mediated pipeline. The caption at stage two is a stored artefact from which all subsequent judgements derive, rather than a post-hoc explanation of a judgement made elsewhere.}
\label{fig:pipeline}
\end{figure}
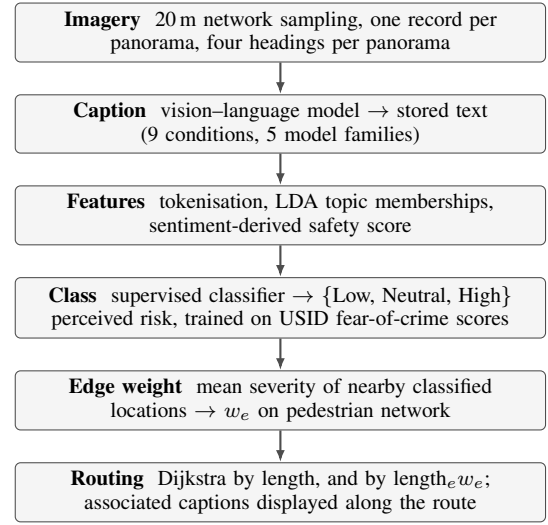

In this section, the approach is presented and justified. The following stages are undertaken: (1) the systematic collection of street-level imagery along the walkable network; (2) the generation and storage of a natural-language caption for each image using a vision--language model; (3) the extraction of interpretable features from that caption text; (4) the supervised classification of each image into a perceived-risk class; (5) the aggregation of classified locations onto pedestrian network edges as a routing weight; and finally, (6) the computation of a distance-optimal and a safety-weighted route over the resulting graph. Fig.~\ref{fig:pipeline} summarises these stages, and each is now presented and discussed in the same order.

\subsection{Image Collection}
\label{subsec:collection}
The study area is defined by the boundaries of the electoral district. The pedestrian and drivable networks are retrieved from OpenStreetMap using OSMnx~\cite{boeing2017osmnx} and then merged, so pedestrian paths and pedestrianised routes absent from the drivable network are also sampled. Candidate locations are interpolated along each edge at a fixed interval $d$, and duplicates are removed. Each candidate is submitted to the Street View Metadata Application Programming Interface (API), and locations that do not return a panorama are discarded. Panorama identifiers are used as the primary key, so that each physical camera position is represented once regardless of how many candidate locations resolve to it. This yielded 61,981 distinct panoramas in Huddersfield and 80,889 in Manchester. For each retained panorama, four images are retrieved at headings $\{0^\circ,90^\circ,180^\circ,270^\circ\}$ with a $90^\circ$ field of view and $0^\circ$ pitch. This configuration is adopted so that the four views span the full horizon at approximately pedestrian eye level with minimal overlap. It is worth noting that Section~\ref{subsec:limitations} reports a defect in the application of this de-duplication step to the Huddersfield collection.

\subsection{Caption Generation}
\label{subsec:captioning}
Each image is converted into a natural-language description using a vision--language model. Because no architecture could be assumed \emph{a priori} to best surface safety-relevant cues, nine conditions across five families were evaluated. More specifically, these were BLIP-base and BLIP-large~\cite{li2022blip}, Florence-2-base and Florence-2-large~\cite{xiao2024florence}, PaliGemma2-3B~\cite{beyer2024paligemma}, Qwen2.5-VL-3B and Qwen2.5-VL-7B~\cite{bai2025qwen25vl}, PaliGemma2-10B, and BLIP-2~\cite{li2023blip2}. Florence-2 was used with task-specific detailed-captioning prompts. Qwen2.5-VL was prompted to describe the scene with attention to perceived safety, thereby testing whether instructing the captioner towards the construct of interest yields more useful text than neutral description. The caption produced at this stage is stored, and it is this stored text, rather than the image, that every subsequent stage operates upon.

\subsection{Supervision and Evaluation Data}
\label{subsec:datasets}
Two independently collected labelled datasets are used throughout. The Urban Security Image Database (USID) provides per-image fear-of-crime ratings from human raters~\cite{guedes2023urban}, and it provides the supervision for the deployed classifier. To the best of the authors' knowledge, it is the largest publicly available dataset giving direct per-image fear-of-crime scores, which is itself indicative of the state of ground truth in this area. However, it contains only 49 images, and Section~\ref{subsec:limitations} treats the consequences of this directly. To strengthen the evaluation, Place Pulse contributes a second independent evaluation set~\cite{salesses2013collaborative,dubey2016deep}. Because its native labels are pairwise comparisons rather than per-image scores, an extreme-groups construction is utilised. This comprises 99 images drawn from the safest and least-safe ends of the perceived-safety ranking, being 50 safe and 49 unsafe, thereby giving a balanced two-class problem.

The two datasets differ in the construct they index, namely fear of crime versus perceived safety, as well as location, capture protocol, and label type. Therefore, every benchmark in Section~\ref{sec:benchmarks} is reported on both and a result is treated as robust only where the two agree. 

The two datasets are not pooled for training. Their labels index different constructs on different scales, which would result in a pooled classifier fitting a blend of the two corresponding to neither. However, pooling remains a reasonable future direction given how small each dataset is individually, and Section~\ref{subsec:transfer} gives grounds to think that training-set size and domain, rather than model capacity, is what currently limits field performance.

\subsection{Caption-Derived Fear-of-Crime Classification}
\label{subsec:classification}
Captions are lower-cased, tokenised, and filtered against standard and domain-specific stop-word lists. The latter removes generic captioning phrasing so that downstream models attend to substantive environmental descriptors, such as \emph{graffiti}, \emph{parked}, \emph{dark}, \emph{alley} and \emph{shops}, rather than to the sentence templates that captioning models tend to reuse.

Then, two families of features are extracted. Latent Dirichlet Allocation (LDA)~\cite{blei2003latent} is fitted over the caption corpus with $K=5$ topics, yielding for each caption a topic-membership vector $\theta_{i1},\dots,\theta_{iK}$ with $\sum_k \theta_{ik}=1$. A small $K$ is chosen, as with 49 supervised images per captioning condition, a larger topic count would leave too few captions to estimate each topic's word distribution stably, thereby favouring sparse topics that shift between re-fits. A sentiment-derived safety score $s_i$ is obtained from a distilled transformer sentiment model~\cite{sanh2019distilbert}, which was selected for its low inference cost over a corpus of this size. The feature vector for image $i$ is therefore $\mathbf{x}_i=[s_i,\theta_{i1},\dots,\theta_{iK}]$. Both feature families are interpretable by construction, and the highest-weighted words per topic are retained so that a given classification can be traced back through topic memberships to the caption text and then to the source image.

Supervision uses the USID dataset. Continuous scores are discretised into three ordered classes using a band of half a standard deviation about the sample mean:
\begin{equation}
g_i=\begin{cases}
\text{low}, & f_i \leq \mu_f - 0.5\sigma_f,\\
\text{neutral}, & \mu_f - 0.5\sigma_f < f_i < \mu_f + 0.5\sigma_f,\\
\text{high}, & f_i \geq \mu_f + 0.5\sigma_f,
\end{cases}
\end{equation}
with $\mu_f\approx3.06$ and $\sigma_f\approx0.86$ for the 49-image sample. Boundaries are defined relative to the empirical distribution rather than to fixed points.

Ten classifiers spanning linear, tree-based, ensemble, margin-based, instance-based and probabilistic families were evaluated. Every corpus-dependent transformation was estimated exclusively from the training portion of each cross-validation fold and applied unchanged to the held-out portion. This covers the count and term-frequency vocabularies, the LDA topic model, the singular value and principal component decompositions, and feature scaling. The sentence and image encoders are pre-trained and are not fitted to this corpus, so they introduce no such dependency. Performance is estimated by stratified cross-validation with the fold count adapted to the smallest class, being $n_{\mathrm{splits}}=\max(2,\min(5,n_{\min}))$. Accuracy, macro-F1 and weighted-F1 are reported together, and macro-F1 is adopted as the primary ranking metric as it is the least sensitive to class imbalance.

\subsection{Safety-Weighted Routing}
\label{subsec:routing}
Classified locations are aggregated onto a pedestrian graph. This graph is built from the OpenStreetMap \texttt{walk} network. Each edge is sampled along its length and matched with classified panorama locations within 50\,m, using a 50\,m spatial grid index in a metric projection. The sampling interval follows the imagery spacing of each collection, being 20\,m in Huddersfield and 100\,m in Manchester, so that in Manchester the samples do not tile the edge continuously. Classes map to severities $\{\text{low},\text{neutral},\text{high}\}\rightarrow\{0,1,2\}$, and for an edge $e$ with non-empty nearby severity set $S_e$ and mean severity $\bar{s}_e$:
\begin{equation}
w_e=\begin{cases}
1+\bar{s}_e, & S_e\neq\emptyset,\\
1.5, & S_e=\emptyset.
\end{cases}
\end{equation}
The additive offset keeps $w_e$ strictly positive, which is necessary because $w_e$ multiplies edge length. A zero weight would render a segment free to traverse irrespective of its length. Edges with no nearby imagery are labelled \emph{Unknown} and assigned 1.5, being the midpoint between \emph{Low} and \emph{Neutral}, rather than being treated as safe by default. These are typically footpaths, given the sparse Street View coverage of pedestrian-only ways. Assigning them the \emph{Low} weight would bias the router towards routing pedestrians along unsurveyed paths precisely because no evidence exists for them, which is the opposite of the intended behaviour. The routing results reported in Section~\ref{sec:routing-behaviour} are not sensitive to this choice. Repeating every condition with unclassified edges weighted at 1.0, which under this mapping is equivalent to \emph{Low} and so leaves them unpenalised, moves the median gain in low-risk route length from 12.78 to 12.02\% on long Manchester journeys and from 7.55 to 7.41 on medium ones. This matters because Manchester has 14.3\% of its network unclassified against 5.2\% in Huddersfield, so a strong dependence upon the \emph{Unknown} weight would have confounded the comparison between the two study areas.

Routing in the deployed application is performed entirely in the client browser over a compact exported graph, utilising Dijkstra's algorithm~\cite{dijkstra1959note} with a binary heap. Two costs are computed per step, namely length, which gives the distance-optimal route, and the product of length and $w_e$, which gives the safety-weighted route. The multiplicative form makes the penalty proportional to the distance actually walked through a segment, which is consistent with $w_e$ representing a condition per metre rather than a fixed entry cost. Routes are reported with distance, estimated walking time at 1.34\,m\,s$^{-1}$, and the proportion of route length in each safety class. Classified locations within 15\,m of the drawn route, measured by the perpendicular distance to the nearest segment, are also used. 

\section{City and Town Deployment}
\label{sec:deployment}

\begin{figure*}[!t]
\centering
\includegraphics[width=\textwidth]{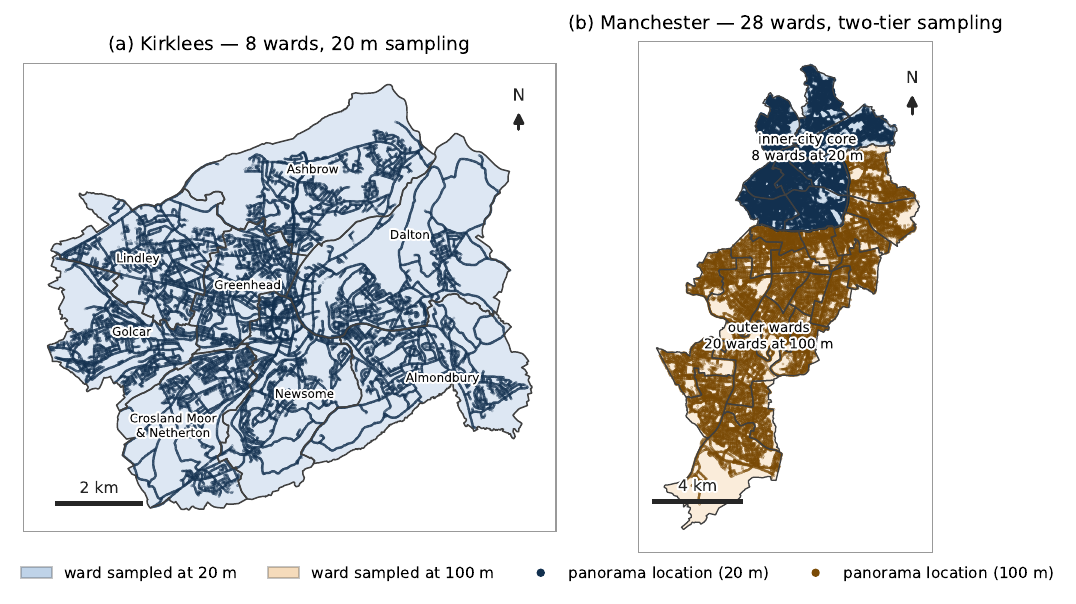}
\caption{Study areas and extraction boundaries. (a)~Huddersfield: eight wards sampled uniformly at 20\,m. (b)~Manchester: 28 wards under a two-tier scheme, with the eight inner-city wards at 20\,m (blue) and the twenty outer wards at 100\,m (orange). Dots are retained panorama locations, so the plotted extent is the actual imagery footprint rather than a nominal administrative boundary; the visibly denser fill of the blue area is the 20\,m regime. Ward polygons are Office for National Statistics 2024 boundaries.}
\label{fig:study-areas}
\end{figure*}

The pipeline was first developed against eight wards covering Huddersfield, in the Kirklees district of West Yorkshire, and was subsequently applied to 28 wards of Manchester city spanning both inner-city and outer suburban areas. Fig.~\ref{fig:study-areas} maps both study areas, and Table~\ref{tab:deployment} summarises their key characteristics.

The two deployments differ in sampling density. Huddersfield was sampled uniformly at 20\,m along the network. Manchester was sampled in two tiers, being the eight inner-city wards (Deansgate, Piccadilly, Ancoats \& Beswick, Miles Platting \& Newton Heath, Ardwick, Harpurhey, Cheetham and Hulme) at 20\,m, and the remaining twenty wards at 100\,m. The coarser interval was adopted for the outer wards in order to keep image-acquisition cost reasonable over an area roughly five times larger than Huddersfield. The inner-city wards retain the finer interval because they carry the densest pedestrian activity. It is worth noting that ward membership of each regime in Fig.~\ref{fig:study-areas} is derived from the collected points themselves rather than from the collection configuration, and that the two regimes separate cleanly by density: 1,247--3,390 locationskm$^{-2}$ in the inner-city wards against 209--973 in the outer wards. 

\begin{table}[!t]
\caption{Deployment statistics for the two study areas.}
\label{tab:deployment}
\centering
\footnotesize
\setlength{\tabcolsep}{4pt}
\begin{tabular}{lrr}
\toprule
& \textbf{Huddersfield} & \textbf{Manchester}\\
\midrule
Electoral wards & 8 & 28\\
Road sampling interval & 20\,m & 20\,m / 100\,m\\
\quad wards at 20\,m & 8 & 8\\
\quad wards at 100\,m & --- & 20\\

Retained locations & 82,589 & 80,967\\
Distinct panoramas & 61,981 & 80,889\\
Images captioned and classified & 330,356 & 323,759\\
Interior locations filtered & 652 (0.8\%) & 1,810 (2.2\%)\\
Pedestrian network nodes & 13,250 & 48,922\\
Pedestrian network edges & 33,846 & 127,368\\
\bottomrule
\end{tabular}
\end{table}

\section{Representation and Captioning Benchmarks}
\label{sec:benchmarks}

\emph{This section addresses RQ1.} A caption-mediated pipeline has two substitutable components that impact upon accuracy, namely the captioning model that produces the text, and the representation that encodes it. Their respective contributions are not known \emph{a priori}. A comparison of captioning models alone, which is the usual form of such an evaluation, holds the representation fixed and therefore cannot separate the two. In this section, both components are evaluated against a common reference. This includes performing the same classification task directly from image embeddings, with no textual intermediate at all.

All conditions pass through an identical downstream analysis. Performance is macro-F1 under $10\times5$-fold stratified cross-validation, reported as mean $\pm$ standard deviation over the ten repetitions, with the best of the ten classifier families selected per condition. Repeated cross-validation is used, as in $n=49$ a single stratified split is unstable. The deployed configuration with topic-and-sentiment features, returns macro-F1 0.638 on a single split but $0.584\pm0.052$ under repetition, with a range of 0.503--0.648 across splits. It establishes that any single-split ranking reported at this sample size should be treated as provisional. The cost of the selection itself is also quantified. Taking the best of the ten classifiers rather than the median inflates macro-F1 by a median of 0.045 in the ten conditions reported here, with a range of 0.012 to 0.093, and six different classifier families are selected across those conditions. It is evident from this that no single family dominates and that the selection affects the absolute level of the reported figures and may also affect close orderings, which is why differences smaller than the observed split and selection variability are read as parity.

\subsection{Direct-Embedding Baseline}
\label{subsec:clip-baseline}
Quantifying the cost of caption mediation requires a comparison with the same classification task performed from image embeddings, with no textual intermediate. In this research, we utilise CLIP ViT-B/32 image embeddings~\cite{radford2021learning}, which are contrastively aligned to language but are not themselves captions. CLIP scores the correspondence between an image and candidate text and cannot generate a description. It therefore serves as a strong direct-image reference for the selected architecture and scale, whilst being unable to supply the auditable artefact that the deployed system requires.

Two variants are reported, being the full 512-dimensional embedding and a six-dimensional principal component analysis (PCA) projection matched to the dimensionality of the deployed caption feature vector. In terms of what this separates, the projection distinguishes the effect of representation \emph{quality} from that of representation \emph{size}.

\subsection{Caption Trade-off}
\label{subsec:ladder}
Table~\ref{tab:ladder} presents a ladder of text representations computed over identical Qwen2.5-VL-7B captions, with the deployed feature encoding and CLIP at its two ends. Row~1 applies the topic-and-sentiment encoding used by the deployed system, and row~5 is the direct image-embedding baseline. Every value is macro-F1 under the repeated cross-validation procedure described above, taking the best of ten classifiers per row. It should be noted that the captioner is the same in rows~1 to~4, so the differences reflect the encoding applied to the text and nothing else. 

\begin{table}[!t]
\caption{Representation over identical Qwen2.5-VL captions.}
\label{tab:ladder}
\centering
\footnotesize
\setlength{\tabcolsep}{4pt}
\begin{tabular}{llcc}
\toprule
& \textbf{Representation} & \textbf{USID} & \textbf{Place Pulse}\\
\midrule
1 & LDA(5) + sentiment & $0.698\pm0.074$ & $0.700\pm0.021$\\
2 & TF-IDF $\rightarrow$ SVD(6) & $\mathbf{0.830\pm0.038}$ & $\mathbf{0.728\pm0.022}$\\
3 & Sentence emb. $\rightarrow$ PCA(6) & $0.742\pm0.041$ & $0.679\pm0.018$\\
4 & Sentence embedding (384-d) & $0.829\pm0.038$ & $0.708\pm0.017$\\
\midrule
5 & CLIP image embedding (512-d) & $0.796\pm0.024$ & $0.696\pm0.030$\\
\bottomrule
\end{tabular}
\end{table}

\begin{figure*}[!t]
\centering
\includegraphics[width=\textwidth]{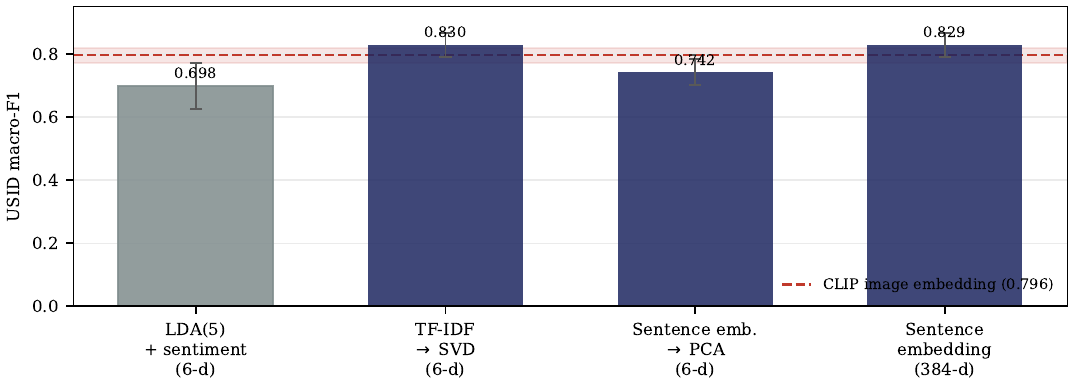}
\caption{Closing the gap to a direct image embedding. Bars show USID macro-F1 under $10\times5$-fold cross-validation over identical Qwen2.5-VL captions; the dashed line and shaded band give the CLIP image-embedding baseline ($0.796\pm0.024$). The grey bar is the topic-and-sentiment encoding used by the deployed system. }
\label{fig:ladder}
\end{figure*}

Row~2 uses the term frequency--inverse document frequency (TF-IDF) weighting reduced by singular value decomposition (SVD), and rows~3 and~4 use sentence embeddings~\cite{reimers2019sentence}. Fig.~\ref{fig:ladder} presents the same comparison graphically. In USID, it is evident that the encoding applied to the text, and not the use of text itself, is what conditions performance. Moving from the six-dimensional topic-and-sentiment vector to TF-IDF reduced to the same six dimensions, over the \emph{same captions}, raises macro-F1 from 0.698 to 0.830. Dimensionality is held constant across that comparison, and so the deficit is not one of capacity. It can be established from this that the topic-and-sentiment encoding discards caption content that simpler term-frequency weighting retains. The best text representation reaches 0.830 against the image embedding's 0.796. That margin of 0.034 is smaller than the standard deviation of the text condition, and the best-of-ten selection inflates every row by a median of 0.045, and so parity with the image embedding is claimed rather than superiority over it. It is noticeable that the encoding which is easiest to interpret directly is also the one that costs the most accuracy. It is worth noting that all caption text is encoded to a maximum of 256 tokens, being the designed sequence length of the sentence encoder utilised, and that this constraint binds for the longer conditions, as 35 of the 49 USID captions exceed it.

Two qualifications attach to this reading. First, the ladder does not rise with representational capacity. Row~3 reaches only 0.742 on USID and 0.679 on Place Pulse, falling below six-dimensional TF-IDF on both, at 0.830 and 0.728, despite using the same sentence embeddings before reduction. Principal component analysis discards more of the signal these captions carry than singular value decomposition over term frequencies does, and so richer text representations are not better in every case here. Second, the Place Pulse column places two text representations above the image embedding, at 0.728 and 0.708 against 0.696. Those margins are smaller than the standard deviations involved, and so parity rather than superiority is claimed. Section~\ref{subsec:transfer} reports that the strongest candidate pipeline identified from these benchmark comparisons did not improve agreement in the field.

\subsection{Captioning Models Under a Fixed Representation}
\label{subsec:vlm-benchmark}
Table~\ref{tab:vlm} provides the captioning-model comparison with the representation held fixed at sentence embeddings, so that differences reflect caption content rather than the encoding applied afterwards. Each figure is macro-F1 under repeated cross-validation with the best of ten classifiers, the final two rows are image-embedding baselines rather than captions, and every condition utilises the instruction with which its caption set was originally generated.

The word counts given in Table~\ref{tab:vlm} are median caption lengths on USID. The Qwen captions were generated using a first-person safety-reasoning prompt and had a median length of 215 words. 

\begin{table}[!t]
\caption{Captioning conditions under a fixed sentence-embedding
representation. Word counts are the median caption length on USID. The two
CLIP rows are image-embedding baselines at 10 and 512 dimensions
respectively.}
\label{tab:vlm}
\centering
\footnotesize
\setlength{\tabcolsep}{4pt}
\begin{tabular}{lccc}
\toprule
\textbf{Condition} & \textbf{Words} & \textbf{USID} & \textbf{Place Pulse}\\
\midrule
Qwen2.5-VL-7B & 247 & $\mathbf{0.829\pm0.038}$ & $0.708\pm0.017$\\
Qwen2.5-VL-3B & 214 & $0.754\pm0.019$ & $\mathbf{0.710\pm0.014}$\\
All captioners (mean emb.) & --- & $0.742\pm0.031$ & $0.661\pm0.023$\\
BLIP-base (10 beams) & 98 & $0.730\pm0.036$ & $0.577\pm0.012$\\
Florence-2-base & 39 & $0.697\pm0.043$ & $0.613\pm0.035$\\
Florence-2-large & 38 & $0.688\pm0.049$ & $0.607\pm0.033$\\
PaliGemma2-10B & 121 & $0.650\pm0.053$ & $0.630\pm0.022$\\
BLIP-large & 15 & $0.642\pm0.048$ & $0.573\pm0.034$\\
BLIP-base & 10 & $0.639\pm0.037$ & $0.566\pm0.017$\\
BLIP-large (10 beams) & 142 & $0.625\pm0.040$ & $0.597\pm0.020$\\
BLIP-2~\cite{li2023blip2} & 10 & $0.591\pm0.072$ & $0.611\pm0.029$\\
PaliGemma2-3B & 11 & $0.564\pm0.046$ & $0.588\pm0.023$\\
\midrule
CLIP CPTED probe & --- & $0.776\pm0.027$ & $0.713\pm0.015$\\
CLIP embedding & --- & $0.796\pm0.024$ & $0.696\pm0.030$\\
\bottomrule
\end{tabular}
\end{table}

As is evident from Table~\ref{tab:vlm}, Qwen2.5-VL-7B leads at 0.829 on USID and Qwen2.5-VL-3B at 0.710 on Place Pulse, both marginally above the respective CLIP baselines of 0.796 and 0.696. However, as the margin is smaller than the standard deviation, it is only possible to state comparable performance. Caption mediation is not any worse at this scale.

\begin{figure*}[!t]
\centering
\includegraphics[width=\textwidth]{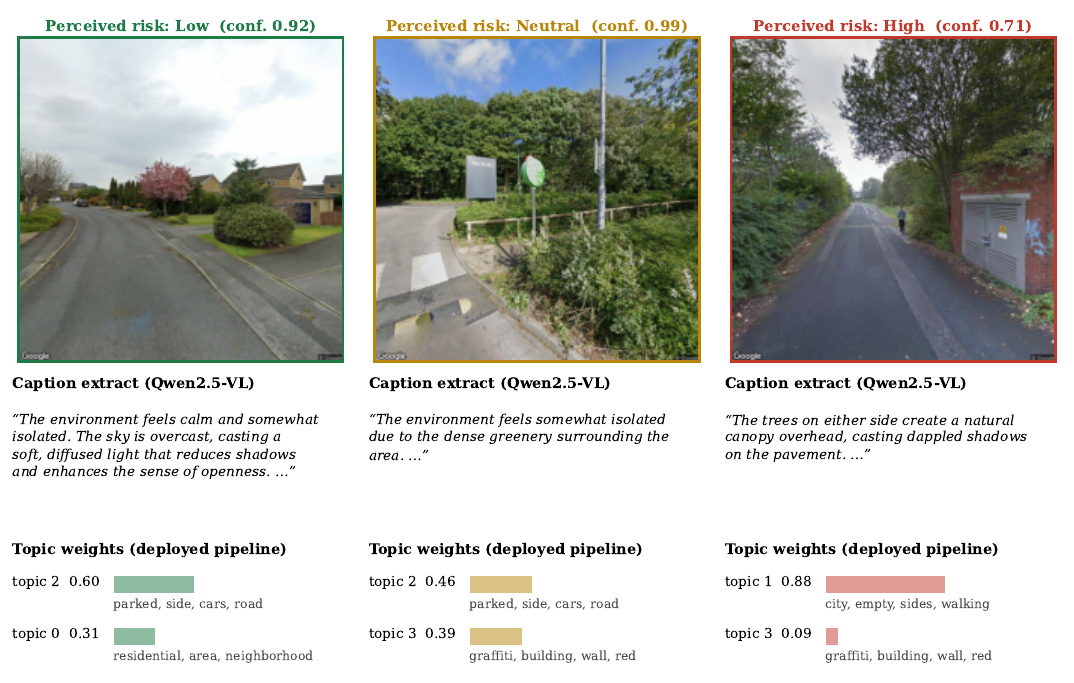}
\caption{One worked example per predicted class, showing the four artefacts
stored for every scored location.}
\label{fig:casestudy}
\end{figure*}

Fig.~\ref{fig:casestudy} presents one worked example per class. The three examples were selected manually from the deployed Huddersfield corpus as readable illustrations of typical street scenes, after excluding interior views that the caption filter had not removed. Their confidences are 0.919, 0.995 and 0.715 for the \emph{Low}, \emph{Neutral} and \emph{High} panels respectively. The field ratings played no part in the selection. The figure is therefore illustrative of the stored artefacts rather than evidence of classifier accuracy. Each panel carries the source image, the caption generated from it, the topic weights extracted from that caption, and the resulting class. A resident or a local authority officer is therefore able to read the basis of a judgement and disagree with it on the evidence. The high-risk example is
instructive in this regard. Its classification is driven by the topic comprising \emph{city}, \emph{empty}, \emph{sides} and \emph{walking} at a weight of 0.883, whereas the graffiti that the caption mentions contributes only 0.092. The model is responding to enclosure and emptiness rather than to the visible disorder, which is a judgement a practitioner is able to accept or reject. No comparable inspection is available for a 512-dimensional embedding.

\subsection{Sensitivity to Design Choices}
\label{subsec:sensitivity}
Two parameters of the deployed encoding were fixed during development and are open to the objection that the reported results depend upon them. Both are therefore varied here, with every other element of the protocol held constant.

Table~\ref{tab:sensitivity} reports the first, being the number of LDA topics. Sweeping $K$ over $\{3,5,10\}$ moves macro-F1 from 0.696 to 0.702 on USID and from 0.685 to 0.700 on Place Pulse. A spread of 0.006 on USID across a threefold change in $K$ establishes that the deployed value of $K=5$ does not affect the finding. At every value of $K$ the topic-and-sentiment encoding remains far below term-frequency weighting at the same six dimensions, and so the deficit reported in Section~\ref{subsec:ladder} belongs to the encoding family rather than to the choice of $K$.

The second is the position of the class boundaries. The deployed classifier cuts the USID fear-of-crime score at 2.62 and 3.48, and the comparison is repeated under three alternatives, being recomputed $\mu\pm0.5\sigma$ cuts at 2.75 and 3.45, tertiles, and quartile-based cuts at the 25th and 75th percentiles. The topic-and-sentiment encoding is last under all four schemes, by margins of between 0.13 and 0.15 macro-F1, and so the central finding is robust to where the lines are drawn.

The same cannot be said of the comparison against the image embedding. Under the deployed cuts and under $\mu\pm0.5\sigma$, term-frequency weighting leads CLIP, at 0.830 against 0.796 and 0.831 against 0.807. Under tertiles and quartiles the ordering reverses, at 0.790 against 0.801 and 0.703 against 0.731. The text representation is therefore at parity with the image embedding rather than above it, and any claim of superiority would not survive a change of cut points. Absolute levels also fall as the cuts move outward, because the neutral class grows from 16 images to 24 whilst the extreme classes shrink.

\begin{table}[!t]
\caption{Sensitivity to topic count and class boundaries.}
\label{tab:sensitivity}
\centering
\footnotesize
\setlength{\tabcolsep}{4pt}
\begin{tabular}{lcccc}
\toprule
& \textbf{LDA+sent.} & \textbf{TF-IDF} & \textbf{Sent. emb.} & \textbf{CLIP}\\
\midrule
\multicolumn{5}{l}{\emph{LDA topic count, USID}}\\
$K=3$      & 0.696 & 0.830 & 0.829 & 0.796\\
$K=5$ (deployed) & 0.698 & 0.830 & 0.829 & 0.796\\
$K=10$      & 0.702 & 0.830 & 0.829 & 0.796\\
\midrule
\multicolumn{5}{l}{\emph{USID class boundaries}}\\
2.62/3.48 (deployed) & 0.698 & \textbf{0.830} & 0.829 & 0.796\\
$\mu\pm0.5\sigma$  & 0.664 & \textbf{0.831} & 0.815 & 0.807\\
Tertiles       & 0.660 & 0.790 & 0.757 & \textbf{0.801}\\
Quartiles (25/75)  & 0.551 & 0.703 & 0.672 & \textbf{0.731}\\
\bottomrule
\end{tabular}
\end{table}

A third design choice concerns the routing cost rather than the classifier. The ordered classes map to severities 0, 1 and 2, which assumes equal spacing between them. The field validation gives reason to doubt that assumption, since the classifier separates \emph{Low} from the remaining classes far more effectively than it separates \emph{Neutral} from \emph{High}. Edge weights were therefore rebuilt from the panorama-level classifications under three alternative monotonic scales, being a convex scale of 0, 1 and 3 that penalises \emph{High} more heavily, a concave scale of 0, 1.5 and 2 that places \emph{Neutral} closer to \emph{High}, and a threshold scale of 0, 0.5 and 3 that treats \emph{Low} and \emph{Neutral} as near-equivalent. A fourth variant replaces the hard class label with a probability-weighted expected severity, which retains the classifier's uncertainty instead of committing to the modal class. Each variant was evaluated over 300 origin--destination pairs per band per city.

Table~\ref{tab:mapping} reports the outcome. The conclusion of Section~\ref{sec:routing-behaviour} is unaffected by the choice of scale. Under all three ordinal alternatives the exchange remains inert below 1\,km and grows monotonically with journey length in both cities, and the long-journey gain in Manchester varies between 7.92 and 9.49\% against 7.68 under the deployed scale. Penalising \emph{High} more heavily returns more low-risk length for a larger detour, which is the expected direction rather than a departure from the reported behaviour.

The probability-weighted variant behaves differently, and its failure is informative. It produces no measurable change in route composition at any journey length, because averaging class probabilities over the panoramas near an edge compresses the weights to a standard deviation of 0.142 against 0.358 under the deployed scale. Almost every edge is assigned a weight near the mean, leaving the router with little to discriminate upon, and 22.3\% of long routes are identical to the distance-optimal path against 5.7\% under the deployed scale. Retaining classification uncertainty in this form therefore collapses distinct predictive distributions onto similar intermediate values rather than preserving the information they carry. This variant is available for Manchester alone, as the Huddersfield prediction records do not retain class probabilities.

\begin{table}[!t]
\caption{Routing outcomes under alternative class-to-cost mappings, Manchester.}
\label{tab:mapping}
\centering
\footnotesize
\setlength{\tabcolsep}{4pt}
\begin{tabular}{lcccc}
\toprule
& \multicolumn{2}{c}{\textbf{Medium (1--3\,km)}} & \multicolumn{2}{c}{\textbf{Long (3--6\,km)}}\\
\cmidrule(lr){2-3}\cmidrule(lr){4-5}
\textbf{Severity scale} & \textbf{Detour} & \textbf{$\Delta$\emph{Low}} & \textbf{Detour} & \textbf{$\Delta$\emph{Low}}\\
\midrule
0, 1, 2 \emph{(deployed)} & $+$0.56\% & $+$1.75 & $+$2.06\% & $+$7.68\\
0, 1, 3 (convex)          & $+$1.07\% & $+$3.01 & $+$3.33\% & $+$9.49\\
0, 1.5, 2 (concave)       & $+$0.58\% & $+$2.02 & $+$2.39\% & $+$8.75\\
0, 0.5, 3 (threshold)     & $+$1.02\% & $+$3.35 & $+$2.98\% & $+$7.92\\
\midrule
Probability-weighted      & $+$0.00\% & $+$0.00 & $+$0.01\% & $+$0.00\\
\bottomrule
\end{tabular}
\end{table}

\section{Routing}
\label{sec:routing-behaviour}

\emph{This section addresses RQ2.} A safety-weighted router is only useful if the weighting changes routes in a way that is worth the detour it costs. The behaviour of the deployed router is therefore characterised independently of the accuracy of the underlying classifier.

The two deployed pedestrian networks differ substantially in composition, and that difference conditions everything reported below. The Huddersfield network comprises 3,520.6\,km of routable length, distributed as 80.6\% \emph{Low}, 13.5\% \emph{Neutral}, 0.7\% \emph{High} and 5.2\% \emph{Unknown}. The Manchester network is larger and markedly less homogeneous, at 9,104.4\,km distributed as 53.3\% \emph{Low}, 30.5\% \emph{Neutral}, 1.8\% \emph{High} and 14.3\% \emph{Unknown}. Huddersfield offers the weighting comparatively little to act upon, since four fifths of that network is already classified \emph{Low}.

Origin--destination pairs were sampled uniformly at random from network nodes and then stratified by the true network length of the distance-optimal route, rather than by straight-line separation. Three bands are used, being short journeys of less than 1\,km, medium journeys of 1 to 3\,km, and long journeys of 3 to 6\,km. The upper band is capped at 6\,km because an unbounded band is dominated by routes in excess of 9\,km, which are not walking journeys. A total of 500 pairs were drawn per band per city, giving 3,000 pairs in total. For each pair, both cost functions were solved to optimality. The deployed browser application uses Dijkstra's algorithm, as described in Section~\ref{subsec:routing}; for the large-scale evaluation reported here, the equivalent optimal routes were computed using A$^\star$ search with an admissible haversine-distance heuristic, in order to reduce computation time over 3,000 pairs. The heuristic remains admissible under the safety-weighted cost because the minimum edge weight is exactly 1.0, so a weighted edge never costs less than its length, and a length never costs less than the straight-line distance it spans. Both procedures therefore return routes of the same optimal cost. Each route was then summarised by its length, its length-weighted mean severity over classified edges, and the proportion of its length in each class.

\begin{table}[!t]
\caption{Routing trade-off by journey length.}
\label{tab:tradeoff}
\centering
\footnotesize
\setlength{\tabcolsep}{4pt}
\begin{tabular}{llrrrr}
\toprule
\textbf{City} & \textbf{Journey} & \textbf{Identical} & \textbf{Detour} & $\Delta$\textbf{\emph{Low}} & $\Delta$\textbf{severity}\\
 & & (\%) & (\%) & (\%) & \\
\midrule
Huddersfield & $<$1\,km & 13.0 & $+$0.00 & $+$0.00 & $+$0.000\\
 & 1--3\,km & 0.8 & $+$0.05 & $+$0.15 & $-$0.001\\
 & 3--6\,km & 0.0 & $+$1.42 & $+$4.27 & $-$0.043\\
\midrule
Manchester & $<$1\,km & 16.0 & $+$0.00 & $+$0.00 & $+$0.000\\
 & 1--3\,km & 7.0 & $+$1.06 & $+$7.55 & $-$0.090\\
 & 3--6\,km & 1.2 & $+$2.73 & $+$12.78 & $-$0.135\\
\bottomrule
\end{tabular}
\end{table}

Table~\ref{tab:tradeoff} reports the result and it is evident that the exchange offered by the weighting is conditioned on the length of the trip. On journeys of under 1\,km the median detour and the median gain in \emph{Low}-risk length are both exactly zero in both cities, and 13.0\% of Huddersfield routes and 16.0\% of Manchester routes are identical under the two cost functions. This suggests that the networks offer fewer practically distinct alternatives over short distances, limiting the effect of the weighting. The effect then grows monotonically with distance. On journeys of 3 to 6 km in Manchester, a median detour of 2.73\% (interquartile range, IQR, 1.15--5.87\%) gains 12.78\% more of the route in \emph{Low}-risk segments (IQR 5.92--22.03\%), and mean severity falls for 97.4\% of pairs, whilst increasing for 0.8\%. It establishes an operating range, whereby the weighting is effective on longer journeys and is close to inert on the shortest ones.

The banded view depends on where the boundaries are drawn, so the relationship is additionally reported as a rank correlation between all 1,500 pairs within each city. Detour correlates with route length at $\rho=+0.443$ in Huddersfield and $\rho=+0.485$ in Manchester, and the gain in \emph{Low}-risk length at $\rho=+0.427$ and $\rho=+0.409$ respectively, each at $p<0.001$. This demonstrates that the dependence upon journey length is a property of the data, rather than an artefact of the bands selected.

The two study areas differ in the magnitude of the exchange and in its shape. On journeys of 3 to 6\,km, Manchester delivers approximately three times the \emph{Low}-risk gain of Huddersfield for roughly twice the detour, which is unsurprising considering how much less homogeneous its network is. However, Manchester has 14.3\% of its network unclassified, compared with 5.2\% in Huddersfield, and this difference requires attention. Unclassified edges carry a weight of 1.5, so part of the difference could reflect avoidance of unsurveyed ground rather than a response to the safety signal. Therefore, every band was recomputed with the unknown weight set to 1.0. The effect is negligible, as quantified in Section~\ref{subsec:routing}. From this it can be established that the difference between cities reflects the safety signal itself.

\begin{figure*}[!t]
\centering
\includegraphics[width=\textwidth]{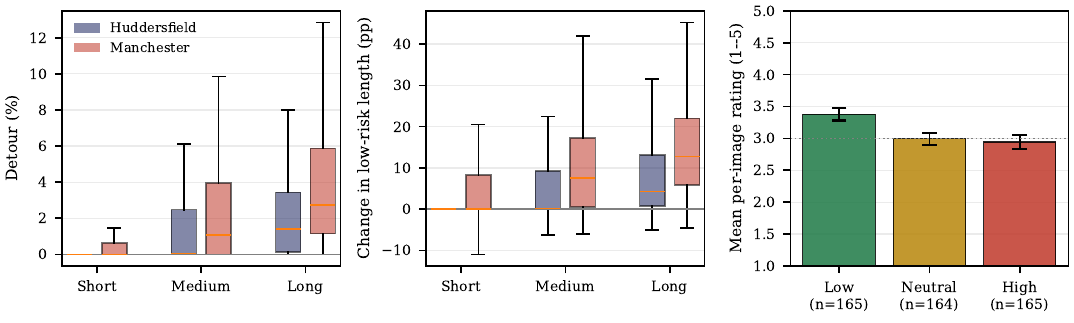}
\caption{Routing behaviour and field agreement. Left and centre: detour
incurred and low-risk length gained by the safety-weighted route, by journey
length and city, over 3,000 origin--destination pairs. Right: mean per-image
field rating by predicted class, with 95\% confidence intervals.}
\label{fig:tradeoff}
\label{fig:byclass}
\end{figure*}

Two qualifications follow. First, \emph{High} segments constitute only 0.7\% of the Huddersfield network and 1.8\% of the Manchester network, so most journeys contain none at all, and the median change in \emph{High}-risk length is zero in five of the six conditions. It is therefore evident that the weighting principally substitutes \emph{Low} for \emph{Neutral} segments rather than avoiding the worst-rated streets. The practical effect of the router is consequently better described as preferring streets that read as reassuring than as avoiding streets that read as dangerous. Second, a small number of sampled routes traverse only unclassified edges, for which mean severity is undefined rather than zero. These number 2 pairs in Huddersfield and 74 in Manchester, and they are excluded from the severity statistics rather than being recorded as unchanged. As is evident in Fig.~\ref{fig:tradeoff}, the trade-off within each band remains graded rather than binary, and so a user who is tolerant of a longer walk obtains a proportionally better route rather than a single fixed alternative.

\section{Validation}
\label{sec:validation}

\emph{This section addresses RQ3.} The benchmark presented above establishes only that caption features track USID ratings, on 49 images collected elsewhere under a different protocol. Whether the deployed classifier tracks the perceptions of people in the deployment area is a separate question. Therefore, an independent rating study was undertaken within the Huddersfield study area.

\subsection{Study Design}
\label{subsec:study-design}
The field ratings were reserved for a single confirmatory use and were not consulted when selecting among captioning models or representations. A sample used to choose between candidates cannot subsequently serve as an unbiased test of the candidate it selected, and the transfer question posed in Section~\ref{subsec:transfer} is answerable only against data that played no part in that selection.\par Rating every classified location was infeasible, and so a two-tier stratified sample was drawn. A fixed \emph{anchor} set of 16 locations is shown to every participant, thereby providing a common basis for inter-rater agreement. Anchors were selected by ranking candidates within each class by classifier confidence and greedily accepting the highest-confidence candidates subject to a 120\,m minimum separation. A \emph{pool} of 480 other locations was stratified by district and predicted class, giving eight districts $\times$ three classes, with an equal allocation of 20 per stratum rather than an allocation proportional to the frequency of the class. This was necessary as the high class accounts for only around 5\% of classified locations, which would result in proportional sampling leaving it critically under-represented. The 120\,m separation threshold, which is six times the camera sampling interval, ensures that sampled scenes are genuinely distinct rather than near-duplicate views.

Participants are shown one image at a time and asked to imagine standing at that location and to rate the degree of safety they feel from 1 (lowest) to 5 (highest). The classifier's prediction is never transmitted to the participant, and so it cannot bias the response. Participation is anonymous in that no personal data is collected, and a browser-generated random token is utilised solely to avoid serving an image twice to the same session. The study received approval from the University of Huddersfield under reference SREP/2026/014, and all participants provided informed consent before submitting ratings. An image-serving endpoint exhausts unseen anchors before drawing from the pool, and then samples at random among the twenty least-rated unseen pool images, which spreads coverage without requiring global coordination between concurrent sessions. Submitted ratings are accepted only if the image was genuinely served to that session, has not already been rated by it, and arrives at least one second after service.

\subsection{Model--Human Agreement}
\label{subsec:agreement}
At the time of analysis, 3,669 ratings had been collected across 494 images from 70 participant sessions, giving a mean of 7.4 ratings per image, a median of 5, and 64 for the anchors. In order to compare an ordinal class against a safety rating, classes were mapped onto the direction of the rating scale, with $\text{low}\rightarrow3$, $\text{neutral}\rightarrow2$ and $\text{high}\rightarrow1$, so that a positive correlation denotes agreement.

The mean ratings ordered as predicted for all three classes, as shown in Fig.~\ref{fig:byclass}, and the correlation between predicted class and rating was statistically significant. Treating each rating as an observation gives Spearman $\rho=0.122$ ($p<10^{-11}$), whereas treating each image's mean as an observation gives $\rho=0.258$ ($p<10^{-8}$). Pearson $r$ gives 0.125 and 0.262 respectively. The per-image figure is the more meaningful of the two, as it is not inflated by those images that happen to have been rated more often.

In absolute terms this is a weak relationship, and the class means are close, being 3.38, 3.00 and 2.94 for \emph{Low}, \emph{Neutral} and \emph{High} respectively. It is evident that the classifier separates \emph{Low} from the other two classes far better than it separates \emph{Neutral} from \emph{High}. 

\subsection{Inter-Rater Disagreement and the Noise Ceiling}
\label{subsec:ceiling}
Human raters disagree substantially with one another on identical images. The mean within-image standard deviation is 1.006 on the five-point scale across all 494 images, and 1.067 on the anchor set, where each image carries approximately 64 ratings. Two raters shown the same photograph differ by 1.17 points on average, and 73.8\% of rater pairs differ by at least one point.

This has a quantifiable consequence on how model performance can be measured. The classifier is evaluated against the \emph{mean} human rating per image; however, that mean is itself a noisy estimate of the underlying population value. By classical attenuation theory, a predictor $X$ correlated with a noisy measure $Y_{\text{obs}}$ of a latent quantity $Y_{\text{true}}$ satisfies
\begin{equation}
r(X,Y_{\text{obs}}) = r(X,Y_{\text{true}})\sqrt{\text{rel}(Y_{\text{obs}})},
\label{eq:attenuation}
\end{equation}
so that even a perfect predictor, with $r(X,Y_{\text{true}})=1$, cannot exceed $\sqrt{\text{rel}(Y_{\text{obs}})}$. That quantity is the \emph{noise ceiling}, being an estimated reliability ceiling for agreement against the present reference under the study's sampling design.

The quantity $\text{rel}(Y_{\text{obs}})$ is estimated by split-half reliability. Each image's raters are randomly partitioned into two equal-sized disjoint halves, per-image means are computed within each half, the two resulting vectors are correlated across images, and the Spearman--Brown correction then projects the half-length correlation to full length~\cite{spearman1910correlation,brown1910some}. Over 2,000 random partitions the mean raw split-half correlation was $r_{1/2}=0.356$, giving $\text{rel}=0.525$ (SD 0.030 across partitions) and a ceiling of
\begin{equation}
\sqrt{0.525}=0.737.
\end{equation}

Against this ceiling, the observed $r=0.262$ represents 35.0\% of the estimated reliability ceiling. Equivalently, the attenuation-corrected correlation is $0.262/0.737=0.355$. It can therefore be established that the classifier is capturing a real but minority share of a signal that is itself only weakly measurable at the current rater density. 

\begin{figure}[!t]
\centering
\includegraphics[width=\columnwidth]{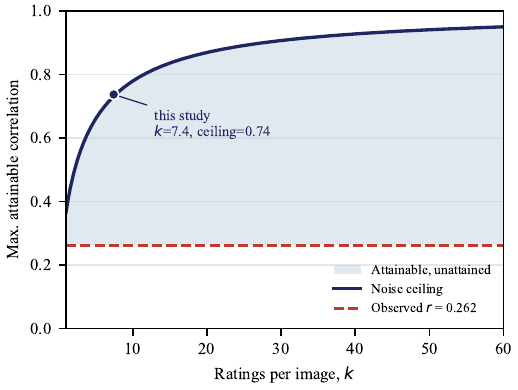}
\caption{Maximum attainable model--human correlation as a function of ratings per image, derived from the estimated single-rating reliability of 0.138. The shaded region is attainable in principle but unattained by the present classifier. At the density of this study ($k=7.4$) the ceiling is 0.737.}
\label{fig:ceiling}
\end{figure}

Back-projecting through Spearman--Brown gives an implied reliability of a \emph{single} rating of 0.138. Fig.~\ref{fig:ceiling} shows the consequence, in that the ceiling rises steeply with rater count, from 0.366 for a single rating to 0.780 at ten ratings per image and 0.870 at twenty. At the rater density typical of field studies, human-reference reliability therefore places a substantial constraint upon the maximum measurable agreement, although the present model also leaves a considerable proportion of the estimated reliability ceiling unexplained.

Table~\ref{tab:validation} provides a consolidated summary of the validation results. The ceiling row gives the maximum correlation any model could attain against this reference at the observed rater density, and the final block reports the single-use confirmatory test of Section~\ref{subsec:transfer}.

\begin{table}[!t]
\caption{Field validation summary.}
\label{tab:validation}
\centering
\footnotesize
\setlength{\tabcolsep}{4pt}
\begin{tabular}{lr}
\toprule
\textbf{Quantity} & \textbf{Value}\\
\midrule
Ratings collected & 3,669\\
Images rated & 494\\
Participant sessions & 70\\
Ratings per image (mean / median) & 7.4 / 5\\
\midrule
Agreement, per rating ($\rho$ / $r$) & 0.122 / 0.125\\
Agreement, per image ($\rho$ / $r$) & 0.258 / 0.262\\
\midrule
Mean within-image SD (1--5 scale) & 1.006\\
Split-half correlation, $r_{1/2}$ & 0.356\\
Reliability after Spearman--Brown & 0.525\\
\textbf{Noise ceiling}, $\sqrt{\text{rel}}$ & \textbf{0.737}\\
Proportion of ceiling attained & 35.0\%\\
Attenuation-corrected $r$ & 0.355\\
Implied single-rating reliability & 0.138\\
\midrule
Candidate pipeline, per image ($r$) & 0.250\\
Deployed vs candidate (Steiger $p$) & 0.840\\
Agreement between the two predictors ($r$) & 0.155\\
\bottomrule
\end{tabular}
\end{table}

Two caveats apply to the ceiling estimate. First, the back-projection to single-rating reliability assumes a common $k$, whereas $k$ varies here from 5 to 65 across images. The estimate should therefore be read as an approximation dominated by the 478 pool images at $k\approx5$. Second, a ceiling computed from split halves of the same participant population bounds agreement with \emph{that} population, and not with some idealised universal perceiver.

\subsection{Confirmatory Transfer Test}
\label{subsec:transfer}
Section~\ref{sec:benchmarks} establishes that a stronger captioner and a stronger text representation together raise benchmark performance substantially, from $0.584$ for the deployed configuration to $0.840$ for the strongest configuration identified. 

Every design choice was therefore fixed before the field data was consulted, namely the captioner (Qwen2.5-VL-7B), the captioning instruction, the representation (sentence embeddings) and the classifier (a support vector machine, SVM, with a radial basis function kernel). Each was selected on USID and Place Pulse alone. The candidate was then trained on the 49 USID images, applied once to all 494 field images, and its predicted classes correlated against the mean human rating. 

The candidate attained $r=+0.250$ (95\% CI $[0.165, 0.330]$) against the deployed system's $r=+0.262$ (95\% CI $[0.169, 0.334]$) on the same images. A Steiger test for dependent correlations sharing a variable~\cite{steiger1980tests} gives $t(491)=+0.202$, $p=0.840$, the statistic being signed so that a positive value favours the deployed system, whereby the difference of $-0.012$ is indistinguishable from zero. A 44\% relative improvement on the supervised benchmark therefore produced no measurable improvement.

Three observations sharpen this result. Firstly, the two systems agree with each other only weakly ($r=0.155$), and so they are not making the same predictions. Two substantially different pipelines arrive at the same accuracy against human ratings, which is what would be expected when a shared external constraint, rather than either model, is binding. Secondly, the ordinal structure is not recovered. Mean human ratings by predicted class were 3.185 (\emph{Low}), 3.208 (\emph{Neutral}) and 2.733 (\emph{High}), and so \emph{Low} and \emph{Neutral} are inverted and statistically indistinguishable, with essentially all discriminative signal lying in the separation of \emph{High} from the remainder. This corroborates, on independent data, the concern raised in Section~\ref{sec:routing-behaviour} regarding the upper two classes. Thirdly, the predicted class distribution collapsed towards \emph{Low} (66.6\%, against balanced training classes), which is consistent with domain shift between USID imagery and UK Street View panoramas.

\subsection{Exploratory Women and Girls Subgroup Analysis}
\label{subsec:subgroup}
Given the VAWG importance and emphasis set out in Section~\ref{sec:intro}, participants were asked one optional question, namely whether they identify as a woman or girl. Of the 70 participant sessions, 69 answered the optional question and one did not. Sessions answering ``yes'' (37 sessions, 2,374 ratings) showed a somewhat stronger per-image correlation than those answering ``no'' or ``prefer not to say'' (32 sessions, 1,263 ratings), being $r=0.261$ against $r=0.138$. The difference did not reach significance under either a Fisher $r$-to-$z$ test ($p=0.087$) or a distribution-free permutation test over 10,000 relabelings ($p=0.081$), which agreed closely. The women and girls subgroup also rated all three classes lower in absolute terms. This is reported as exploratory as the comparison group is small. 

\section{Discussion}
\label{sec:discussion}

The architecture proposed in this paper stores a natural-language description of every scored location and derives the perceived-risk class entirely from structured features of that stored text. The benchmark analysis indicates that this design costs little in accuracy. Against a CLIP baseline of macro-F1 of 0.796 on USID, text representations over identical captions reach 0.830, and on Place Pulse they reach 0.728 against 0.696. Parity with direct image embedding is claimed rather than superiority, on two grounds. First, the margins are smaller than the standard deviations of the text conditions, and on USID the ordering reverses under two of the four class-boundary schemes examined in Section~\ref{subsec:sensitivity}. Parity is nonetheless the substantive result, against an expectation that inserting a textual intermediate carries a material accuracy penalty.

The topic-and-sentiment encoding reaches 0.698 over the same captions that support 0.830 under term-frequency weighting at identical dimensionality. The deficit is therefore a property of the encoding applied to the text, and not of the decision to use text. The distinction has practical value, as an encoding can be replaced without regenerating any captions, whereas substituting the captioning model requires the corpus to be processed again. Section~\ref{subsec:sensitivity} demonstrates the finding to be robust to the number of topics, which moves macro-F1 by 0.006 across $K\in\{3,5,10\}$, and to the position of the class boundaries, under all four of which the topic-and-sentiment encoding ranks last.

Accuracy alone does not capture what the stored caption provides. A ten-dimensional probe of CLIP constructed from named environmental-design constructs recovers 97\% of the opaque embedding performance, so interpretable features are not intrinsically weak, and interpretability alone does not distinguish the proposed architecture. What distinguishes it is legibility. People can inspect the generated description on which the classification is based, compare it with the source image, and contest the resulting judgement, which is not available with a vector of construct scores nor for a 512-dimensional embedding. 

The routing application establishes that a stored safety score estimate of this kind supports a usable transport function. The safety weighting produces a graded exchange rather than a binary alternative, and its magnitude is conditional upon journey length. On journeys of 3 to 6\,km a median detour of 2.73\% returns 12.78\% more low-risk length in Manchester, and 1.42\% returns 4.27\% in Huddersfield. Below 1\,km the weighting has little effect in either city, as few alternative paths exist over such distances. These results establish an operating range in which the weighting is effective on longer journeys but largely inert on shorter ones.

The deployment demonstrates the technical portability of the pipeline, although perceptual validity was evaluated only in Huddersfield and cannot yet be assumed to transfer to Manchester. The pipeline was developed against eight Huddersfield wards and subsequently applied to 28 Manchester wards without retraining and without alteration of the method, covering 654,115 images across 36 electoral wards. The two networks differ substantially in composition, with \emph{Low}-classified segments accounting for 80.6\% of routable length in Huddersfield and 53.3\% in Manchester, and the weighting behaves consistently across both. The approach is therefore a reusable procedure rather than a configuration fitted to one location.

What the router can achieve is bounded by the scoring of images. With \emph{High}-classified segments accounting for 0.7\% and 1.8\% of routable length in the two study areas, it mainly selects between \emph{Low} and \emph{Neutral} segments rather than avoiding the high-risk ones. Whether the scarcity reflects the study areas or a conservative classifier is unresolved. The field validation is consistent with the second reading, as the classifier separated \emph{Low} from the remaining classes more effectively than it separated \emph{Neutral} from \emph{High}, with class means of 3.38, 3.00 and 2.94. Distinguishing the two requires in-domain labelled data.

Independent validation against local perception places the benchmark results in context. Agreement with raters in the field study is statistically significant but modest, at $r=0.262$ per image, against a ceiling of 0.737 imposed by disagreement among the raters themselves. Benchmark improvement and deployment improvement were unrelated in this case, as reported in Section~\ref{subsec:transfer}. A plausible explanation is the substantial domain mismatch: USID comprises 49 images collected under a different protocol, in a different country, and with a different camera geometry from the panoramas the system scores, so selection over it optimises fit to those images rather than to the deployment. 

The work has identified that the result depends on the way agreement is reported. Correlations between a model and a human reference are bounded by the reliability of that reference, so two systems of identical quality validated against references of differing rater density will report different correlations. Three practices follow: report the number of ratings per item and the observed dispersion; report an estimated ceiling and the proportion of it attained; and prefer attenuation-corrected agreement when references differ in reliability. The third is not neutral with respect to the present results, as the headline figure rises from 0.262 to 0.355 under correction, which is why the uncorrected value, the ceiling, and the estimation method are all reported.

Several design decisions are based on the potential for crime to occur rather than on performance considerations. Outputs are presented as indicators of perceived safety derived from visible environmental cues, and not as predictions of crime. Results are reported at segment level rather than as neighbourhood rankings, since ranking areas invites deficit framing of places and populations already subject to it. Unclassified edges are weighted at 1.5 rather than treated as safe, at some cost in routing efficiency, so that the absence of evidence is not converted into evidence of safety. 

One risk appears specific to safety-aware routing and, to the best of the authors' knowledge, is not addressed in the existing literature. Systems of this kind divert pedestrian traffic away from segments scored as less safe. Reduced footfall reduces natural surveillance, which is one of the conditions that make a street feel safer and, according to established accounts of urban vitality, safer~\cite{jacobs1961death,newman1972defensible}. A widely adopted perceived-safety router could therefore alter the quantity it measures, through a feedback loop that a cross-sectional evaluation cannot observe, including the present one. This supports treating such systems as instruments for identifying where environmental intervention is warranted, rather than solely as tools for individual avoidance.

\subsection{Limitations}
\label{subsec:limitations}
The most substantial limitation concerns the supervised benchmark, which rests on 49 images. Three-class cross-validation at that scale yields wide confidence intervals, makes the ranking in Table~\ref{tab:vlm} unstable to individual images, and cannot support strong claims regarding relative captioning model quality.  The consequences of the sample size are observable: Section~\ref{subsec:transfer} reports that selection over that benchmark yielded an apparent improvement of 44\% that did not transfer to field data. The field validation of Section~\ref{sec:validation} was conducted because the extrapolation from 49 images to 654,115 requires independent checking, and it returned significant but modest agreement.

A second group of limitations bounds the interpretation of the benchmarks. Per-condition figures select the best of ten classifier families, which is optimistic in level by a median of 0.045 macro-F1. The inflation is not uniform, ranging from 0.027 for the CLIP baseline on USID to 0.093 for a caption representation on the same dataset, so the caption-to-embedding comparison in Table~\ref{tab:ladder} is understated rather than overstated by the procedure. Classifier-family selection and performance estimation used the same repeated cross-validation procedure, so the reported values are comparative estimates rather than unbiased estimates of generalisation performance. The CLIP baseline uses one architecture at one scale, being ViT-B/32, and is not an upper bound upon embedding-based methods generally. The CPTED probe prompts were authored by the researchers with reference to the theoretical literature and were not validated against independent coders, and the probe is reported as an exploratory proof-of-concept upon which no deployment decision rests. The confirmatory test compares two complete pipelines rather than isolating any single component, which follows from reserving the field ratings for a single use; it establishes that the candidate did not improve upon the deployed system, and not which component was responsible. Section~\ref{subsec:future} sets out the component-level comparisons that a fresh sample would permit.

A third group concerns the deployed system and the scope of its evaluation. The safety weighting is a static snapshot and does not vary with time of day, season or lighting conditions, although perceived safety does, which is a substantive gap for a system whose motivating use case includes evening journeys. The interior-image filter is a caption keyword heuristic rather than a trained classifier, and its effect is bounded by the proportion of retained locations upon which it acts, being 0.8\% in Huddersfield and 2.2\% in Manchester. Within the router, nearest-node snapping uses a linear scan, which is adequate at study-area scale but would require a spatial index at national scale. The validation study covers one city, and self-selected volunteers recruited through the project website are not a representative sample of the pedestrian population, so the ceiling reported in Section~\ref{subsec:ceiling} bounds agreement with that population rather than with an idealised universal perceiver. Two properties of the caption representation also warrant statement. The sentiment score is produced by a general-purpose model trained on consumer review text, and linguistic sentiment is a related but distinct construct from perceived safety. Terms such as \emph{quiet}, \emph{crowded} and \emph{dense vegetation} carry a sentiment polarity that need not match their implication for a pedestrian's sense of safety, and the score was not validated against safety judgements in this domain. The topic representation is a bag-of-words model, so it preserves neither word order nor the scope of negation: the captions \emph{vegetation blocks visibility along the footpath} and \emph{vegetation does not block visibility along the footpath} contain almost the same terms and receive similar topic memberships despite opposite meanings. Section~\ref{subsec:sensitivity} shows that replacing the topic model with term-frequency weighting at identical dimensionality raises macro-F1 from 0.698 to 0.830, which establishes that the topic representation discards a substantial amount. It does not isolate the loss attributable to negation specifically, since term-frequency weighting is equally order-blind, and the sentence-embedding condition, which does encode order, reaches 0.829 rather than exceeding it. Finally, the approach estimates \emph{perceived} safety from visible environmental cues. It is not a predictor of crime, and outputs are presented to users with that distinction made explicit.

\subsection{Future Work}
\label{subsec:future}
In terms of validation, Fig.~\ref{fig:ceiling} motivates a study designed for depth rather than breadth. More specifically, collecting 20 to 30 ratings per image on a smaller image set would raise the estimated ceiling to approximately 0.87--0.91, and would permit a sharper estimate of model quality than the present design allows. Repeating the rating study in the second deployment city would establish whether classification quality transfers, and not only the pipeline that produces it.

The ratings collected here constitute the beginnings of a ground-truth dataset for the deployment area. Retraining or recalibrating against them, rather than against 49 images collected elsewhere, follows from Section~\ref{subsec:transfer}, which localises the limitation to the supervision data rather than to the captioner or the representation. The immediate diagnostic is cross-validation \emph{within} the field-rated images. If in-domain training lifts agreement materially above 0.25, the binding constraint is the out-of-domain supervision and the remedy is to collect further field ratings. If it does not, the constraint lies closer to the reliability of the construct itself. It is worth noting that this analysis requires no further imagery.

Two comparisons follow, and both require a fresh independent sample rather than the ratings already spent. Re-encoding the stored captions is computationally inexpensive and requires no re-captioning, whereas substituting the captioner does. A positive result from the first would implicate the representation in the benchmark-to-deployment gap, whereas a null result would place the limitation in the supervision. The immediate priorities are therefore in-domain recalibration, validation in Manchester, and the development of time-varying route weights.

\section{Conclusion}
\label{sec:conclusion}

This paper has presented a pedestrian routing system in which perceived safety is estimated through an explicit, stored, natural-language intermediate representation. Nine captioning conditions and a direct image-embedding baseline were benchmarked under an identical downstream pipeline, and the resulting approach was deployed over 654,115 street-level images across 36 wards in two study areas, transferring between them without retraining. Addressing the questions set out in Section~\ref{sec:intro}, caption mediation is not the accuracy bottleneck it appears to be, as holding the captioner fixed and changing only the text representation raises macro-F1 from 0.698 to 0.830 against a CLIP baseline of 0.796, so parity with image embeddings is claimed rather than superiority over them (RQ1). The safety weighting offers a measurable, length-conditional exchange, returning 12.78\% more low-risk route length for a median detour of 2.73\% on journeys of 3 to 6\,km whilst having little effect below 1\,km (RQ2). Agreement with raters in the field study is statistically significant but modest, at $r=0.262$ per image, is bounded at 0.737 by disagreement among those residents themselves, and was not improved by a pipeline 44\% stronger on the supervised benchmark ($r=0.250$, $p=0.84$) (RQ3).

Two findings extend beyond the system described here.  The first is that benchmark gains did not predict field gains, which was observable only because the field ratings were reserved for a single confirmatory use. The second is that agreement statistics in this area are not comparable across studies unless the reliability of the human reference is reported alongside them. Inter-rater disagreement bounds attainable correlation at 0.737 in the data collected here, and the reliability of a single rating is 0.138, so reporting rater density, inter-rater dispersion and an estimated ceiling alongside agreement is recommended.

An architecture that stores a readable intermediate therefore costs little in measured accuracy whilst producing a segment score that a resident or a practitioner is able to read and contest. The results indicate that collecting in-domain labelled data is a more promising next step than further selection among the captioning models and representations evaluated here, and collecting such data at scale, alongside time-varying safety weightings, is the direction these results indicate. Throughout, the intention of such capabilities is to complement the qualitative expertise of practitioners working in environmental design and community safety, and not to serve as a replacement for it.

\bibliographystyle{IEEEtran}
\bibliography{refs}

\begin{IEEEbiography}
[{\includegraphics[width=1in,height=1.25in,clip,keepaspectratio]{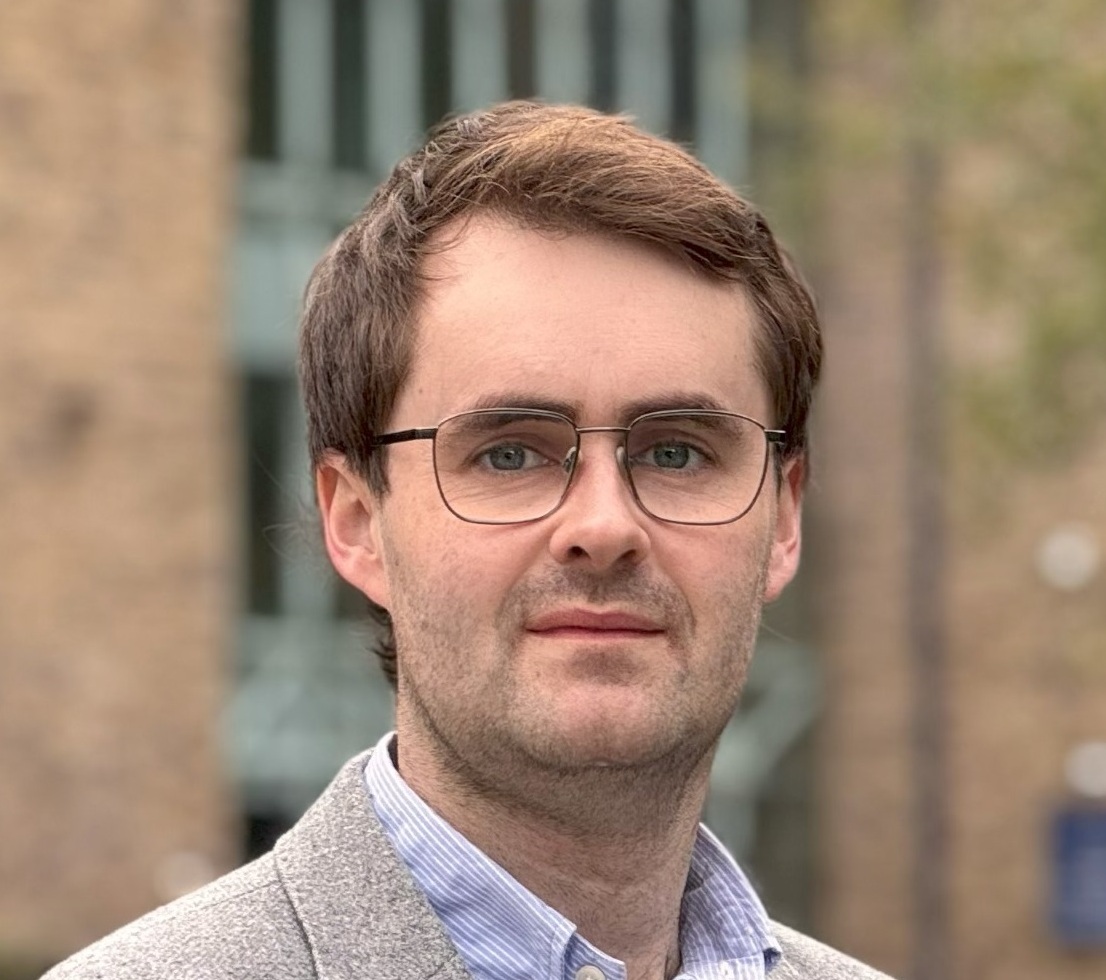}}]{Simon Parkinson}
is Professor of Cyber Security with the School of Computing and Mathematics at Manchester Metropolitan University. His research interests include digital identity and access control, the application of artificial intelligence to security analysis and risk assessment, and automated planning. In addition to his research in cyber security, he has a sustained interest in the computational analysis of the built environment, including automated Crime Prevention Through Environmental Design assessment and the environmental correlates of fear of crime. Prof. Parkinson is a member of the UK Government's Cyber Security Advisory
Board.
\end{IEEEbiography}
\vskip -2\baselineskip plus -1fil

\begin{IEEEbiography}
[{\includegraphics[width=1in,height=1.25in,clip,keepaspectratio]{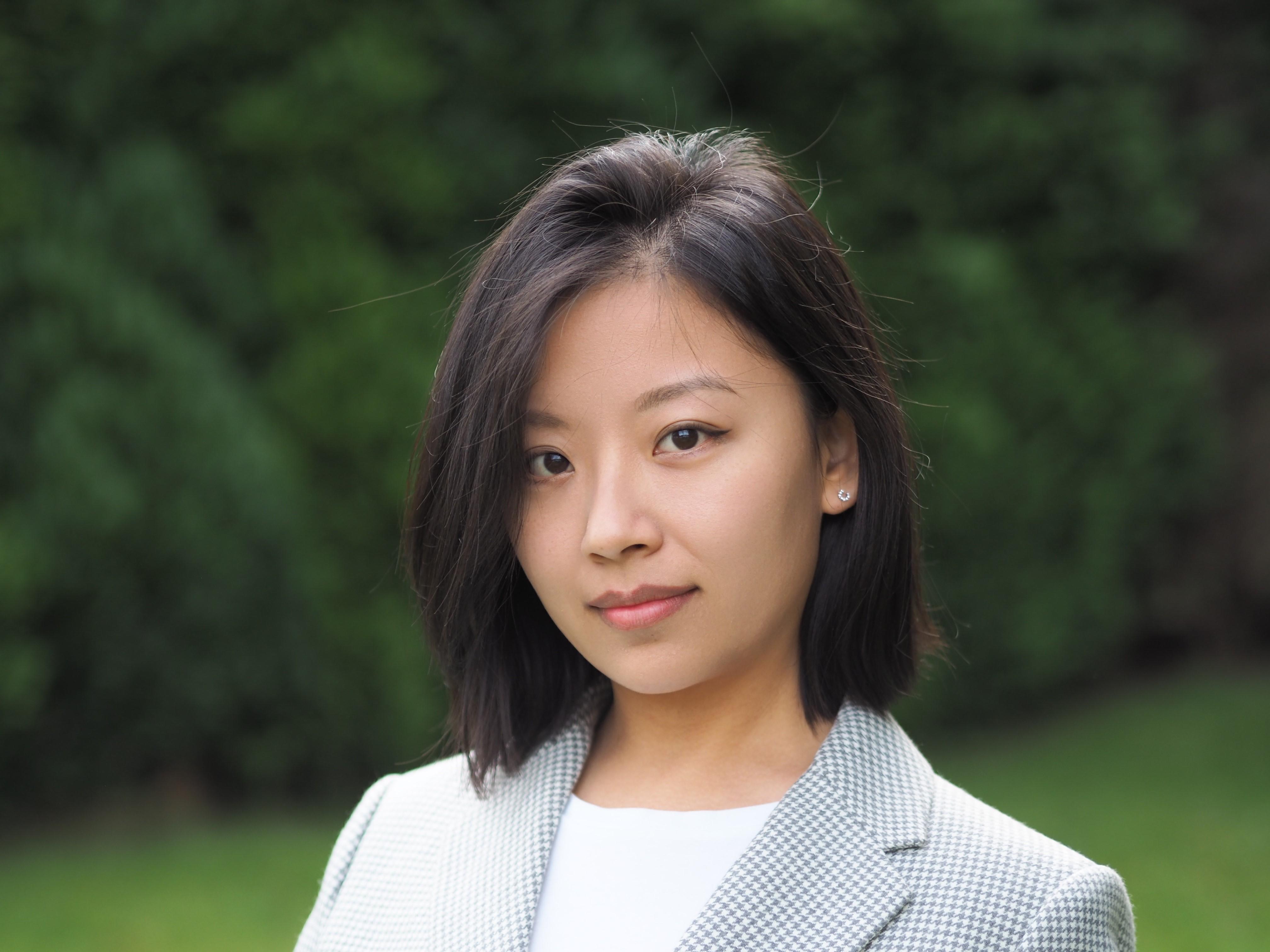}}]{Paloma Liu}
is a Senior Lecturer in Project Management with Huddersfield Business School at the University of Huddersfield. Her research interests include business analytics, operations and supply chain management, and the application of artificial intelligence and data-driven methods to decision-making. She also has an interest in sustainable mobility and the analysis of the built environment, particularly pedestrian experiences and perceived safety. Dr Liu is a Fellow of the Higher Education Academy and a certified Project Management Professional.
\end{IEEEbiography}
\vskip -2\baselineskip plus -1fil

\begin{IEEEbiography}
[{\includegraphics[width=1in,height=1.25in,clip,keepaspectratio]{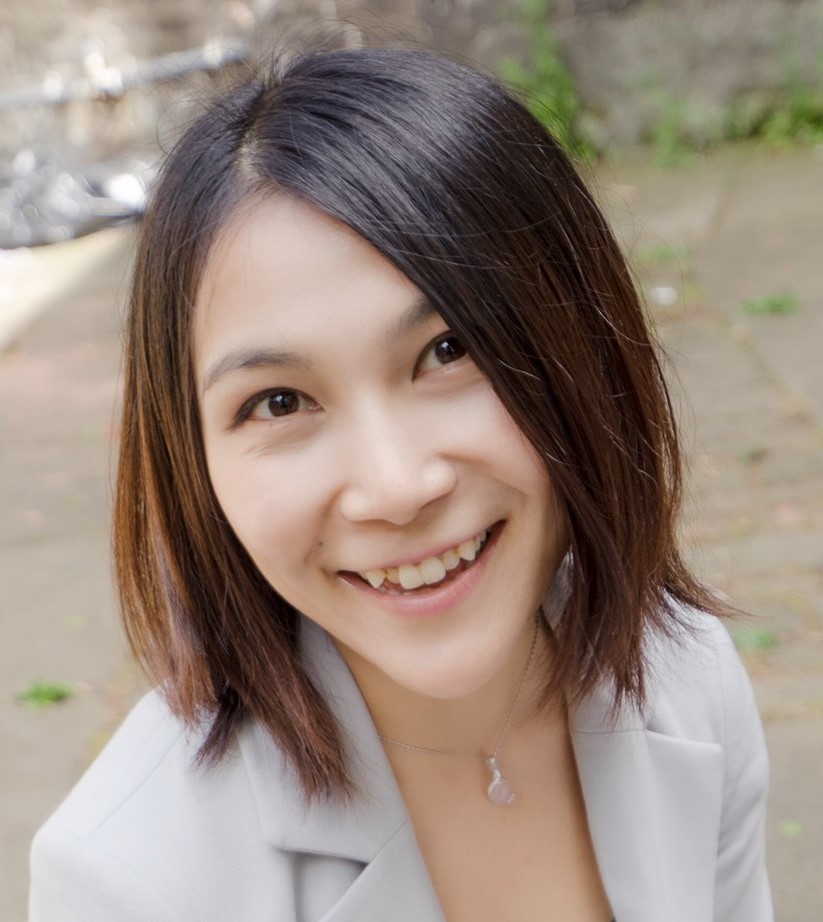}}]{Wei Zheng}
is a Senior Lecturer in Spatial Planning at the University of Manchester. Her research examines spatial inequality and how it is shaped by urban transformation, infrastructure, human behaviour and policy decisions. She uses spatial data analysis, geospatial methods and AI-enabled approaches to identify and interpret inequalities across places and population groups. Her empirical work focuses on health, mobility and low-carbon transitions, with a broader interest in how planning can address uneven urban development and support more equitable policy interventions.
\end{IEEEbiography}
\vskip -2\baselineskip plus -1fil

\begin{IEEEbiography}
[{\includegraphics[width=1in,height=1.25in,clip,keepaspectratio]{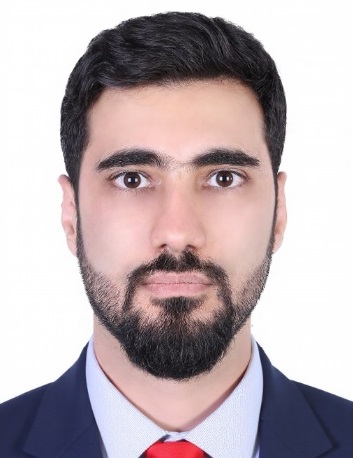}}]{Mohammadreza Sheikhfathollahi} received the B.Sc. degree in Electrical Engineering from Islamic Azad University of Shahr-Rey, Tehran, Iran, in 2017, and the M.Sc. degree in Telecommunication Engineering from Islamic Azad University Science and Research Branch, Tehran, Iran, in 2019. He is now pursuing the Ph.D. degree in Computer Science at the University of Huddersfield, UK (2024–present). His research interests include Machine Learning, Pattern Recognition, Image Processing, and Large Language Models (LLMs), with applications in intelligent surveillance systems, and multimodal data analysis. 

\end{IEEEbiography}
\vskip -2\baselineskip plus -1fil

\end{document}